\documentclass[conference]{IEEEtran}
\usepackage{cite}
\usepackage{amsmath,amssymb,amsfonts}
\usepackage{algorithmic}
\usepackage{graphicx}
\usepackage{textcomp}
\usepackage{xcolor}
\usepackage{subcaption}
\usepackage[export]{adjustbox}
\usepackage{multirow}
\def\BibTeX{{\rm B\kern-.05em{\sc i\kern-.025em b}\kern-.08em
    T\kern-.1667em\lower.7ex\hbox{E}\kern-.125emX}}

\begin{document}

\title{Activation Analysis of a Byte-Based Deep Neural Network for Malware Classification}

\author{\IEEEauthorblockN{Scott E. Coull}
\IEEEauthorblockA{\textit{FireEye, Inc.} \\
scott.coull@fireeye.com}
\and
\IEEEauthorblockN{Christopher Gardner}
\IEEEauthorblockA{\textit{FireEye, Inc.} \\
christopher.gardner@fireeye.com}
}

\maketitle

\begin{abstract}
Feature engineering is one of the most costly aspects of developing effective machine learning models, and that cost is even greater in specialized problem domains, like malware classification, where expert skills are necessary to identify useful features.  Recent work, however, has shown that deep learning models can be used to automatically learn feature representations directly from the raw, unstructured bytes of the binaries themselves.  In this paper, we explore what these models are learning about malware. To do so, we examine the learned features at multiple levels of resolution, from individual byte embeddings to end-to-end analysis of the model.  At each step, we connect these byte-oriented activations to their original semantics through parsing and disassembly of the binary to arrive at human-understandable features.  Through our results, we identify several interesting features learned by the model and their connection to manually-derived features typically used by traditional machine learning models.  Additionally, we explore the impact of training data volume and regularization on the quality of the learned features and the efficacy of the classifiers, revealing the somewhat paradoxical insight that better generalization does not necessarily result in better performance for byte-based malware classifiers.
\end{abstract}

%\begin{IEEEkeywords}
%component, formatting, style, styling, insert
%\end{IEEEkeywords}

\section{Introduction}
\label{sec:intro}
To effectively protect users from the latest malware threats, detection mechanisms must be capable of adapting as quickly as the threats themselves. Traditional machine learning-based antivirus (i.e., next-gen AV) solutions provide this capability by generalizing from previous examples of malware, but often require laborious development of hand-engineered features by domain experts to gain a true advantage. Moreover, these features are often specific to each type of executable file (e.g., Portable Executable, Mach-O, ELF, etc.), further compounding the amount of overhead required. Recently, however, a series of deep neural network models \cite{johns2017cnn, raff2017malware, krvcal2018deep} have been proposed that operate directly on the raw bytes of executable files to detect malware - effectively learning the feature representations directly from the data with no information about its syntax or semantics.  Surprisingly, these byte-based classifiers have reported accuracy exceeding 0.96 and area under the ROC curve (AUC) of greater than 0.98.
 
Given the success of these approaches, an obvious question arises: what exactly are these neural networks learning? Answering this question is important in developing a rigorous understanding of the generalization capabilities of these models, as well as their robustness to evasion.  Furthermore, this analysis will help to shed light on the connection between manual feature engineering and feature representation learning in the malware classification domain. In this paper, we seek to answer this question by providing a deep and broad analysis of activations in a byte-based deep neural network classifier that is representative of the architectures proposed in previous work. Unlike previous work, however, we expand our analysis beyond simply looking at the location of the activations to understanding the specific features that are learned and their connection to the semantics of the executable as a malware analyst would understand them.  We perform this analysis under a variety of training regimes to gain a better understanding of the bias-variance tradeoff that exists for byte-based models in the unique problem area of malware classification.

Specifically, we examine the question at three levels: (1) the embedding layer to uncover learned similarities among independent byte values,  (2) the first convolutional layer to identify low-level features over short byte sequences, and (3) end-to-end analysis for complex features combined over several layers of aggregation in the model. At each of these layers, we compare three models trained under increasing data volumes and levels of regularization to understand the relationship between these training variables, the features learned by the models, and the efficacy of those models in correctly classifying malware. Where possible, we bridge the gap between raw-byte activations and the semantics of the executable through automated parsing and disassembly of the activation locations in an effort to obtain human-understandable explanations for the model's predictions.

Overall, the results of our experiments surface a number of interesting findings about the use of byte-based deep neural networks for malware classification.  For one, our results highlight the importance of depth in learning rich, semantically-meaningful features.  Simple code-related features only appear as important features at the lowest levels of the models, while end-to-end features tended to mirror those features typically derived from manual feature engineering efforts (e.g., invalid checksum, presence of certificates in signed binaries, etc.).  Meanwhile, import-related features (i.e., Windows APIs) are present throughout all levels of our analysis, from the embedding layer to end-to-end features.  Perhaps the most important of these findings is a paradoxical one in which increased training data volume and regularization results in more generic features applicable to both classes, but decreased classification efficacy -- perhaps indicating a degree of bias toward malware-oriented features is beneficial in malware classification.

The remainder of the paper is organized as follows.  We begin in Section \ref{sec:background} by providing background about the representative architecture we examine, the data used to train and evaluate the models, and the tools used in our experiments.  Next, in Sections \ref{sec:embedding} through \ref{sec:end-to-end}, we compare our findings for the embedding, low-level features, and end-to-end features from three models trained with increasing levels of data volume and regularization.  Finally, in Section \ref{sec:conclusion}, we provide discussion of the results and their implications for future development on byte-based deep neural network models for malware classification.

\section{Background}
\label{sec:background}
In this section, we begin by first discussing the primary components of our analysis and the overall methodology we will use to extract human-understandable features from the model.

\begin{figure}[t]
\centering
\includegraphics[width=\columnwidth]{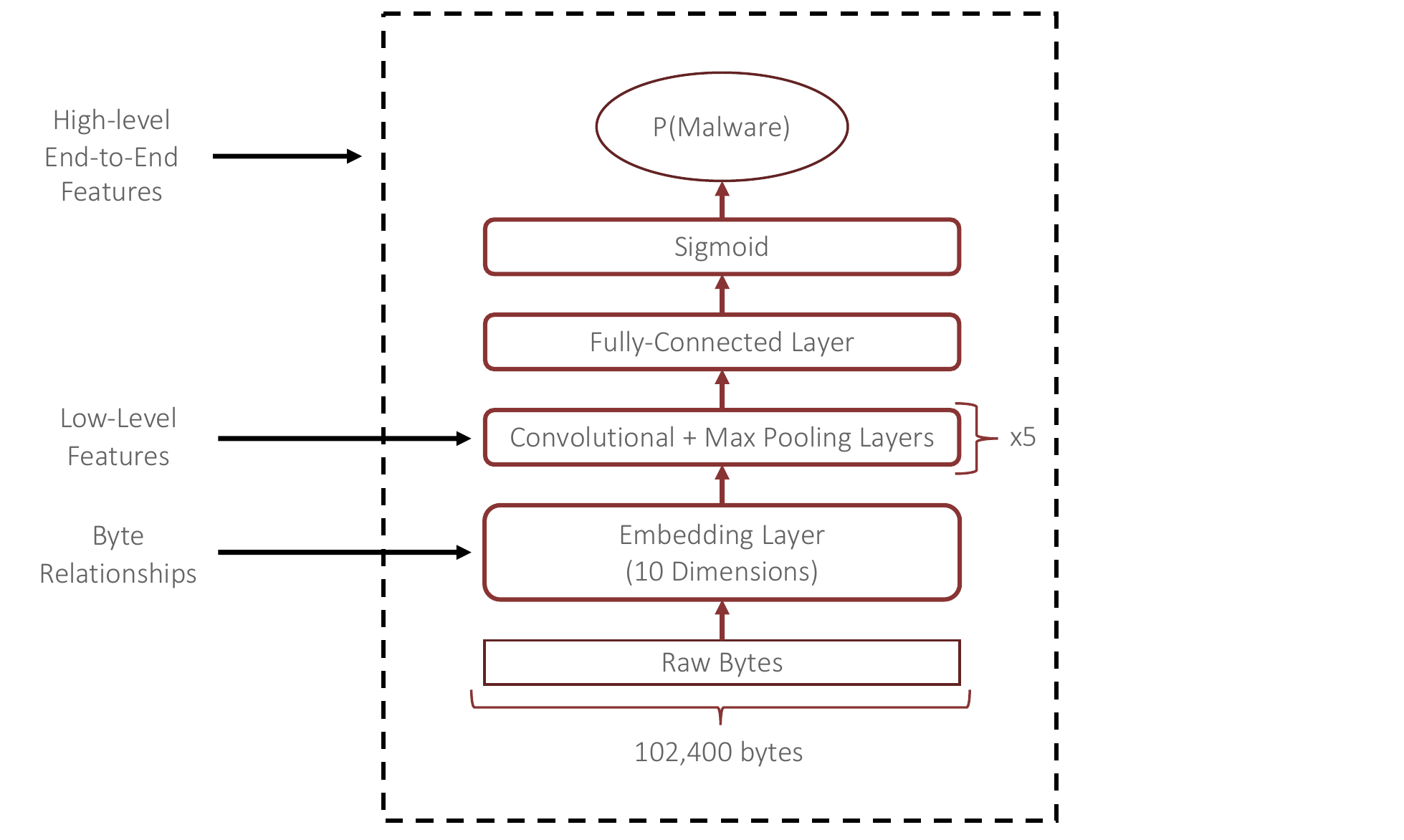}
\caption{High-level CNN architecture.}
\label{fig:arch}
\end{figure}

\subsection{Model Architecture}
\label{sec:background:arch}
Previous work on byte-based malware classifiers has been heavily influenced by convolutional neural network architectures (CNNs) for character-level text classification proposed by Zhang et al. \cite{zhang2015charcnn}.  In general, previous models use a CNN architecture over a fixed-length byte sequence with an embedding layer and varying numbers of convolutional and pooling layers.  The MalConv architecture proposed by Raff et al. \cite{raff2017malware}, for instance, uses a single, gated convolutional layer and global pooling to accommodate large input sequences, while Kr{\v c}{\' a}l et al. \cite{krvcal2018deep} uses four convolutional layers and Johns \cite{johns2017cnn} uses five.  All architectures except for  Kr{\v c}{\' a}l et al. use a learnable embedding layer.  In addition, all architectures treat the input byte sequence equally without taking into account contextual information about structure, syntax, or semantics.

The architecture that we use in this paper, as shown in Figure \ref{fig:arch}, is essentially the same as the one proposed by Johns \cite{johns2017cnn}.  It contains a learnable, 10-dimensional embedding layer and five alternating convolutional and pooling layers that hierarchically combine feature outputs from previous layers.  A single fully-connected layer and sigmoid function are used to perform classification.  The input length of our classifier is restricted to a 100KB sequence, where longer sequences are truncated and shorter sequences are padded with a distinguished padding symbol (i.e., the input alphabet contains 257 symbols).  We note that examining the relationship between feature representation and model architecture remains an interesting avenue of future work, though due to space constraints could not be experimentally pursued here.  Interested readers can find a comparison of our findings with those of Demetrio et al. on the MalConv architecture \cite{luca2019explaining} in Section \ref{sec:conclusion}.

\subsection{Data}
\label{sec:background:data}
In our experiments, we use separate datasets for training, efficacy testing, and activation analysis.  Our \textit{baseline} dataset consists of 15.62M distinct Windows Portable Executable (PE) files \cite{windowspe} collected between July 2015 and July 2017 from a combination of VirusTotal \cite{virustotal}, ReversingLabs \cite{reversinglabs}, and other proprietary sources. The dataset was downsampled using a stratified sampling strategy from a collection of more than 100M binaries to ensure uniform representation across malware families, and the final dataset contained 20\% malware and 80\% goodware.  Additionally, we created a \textit{small} dataset that is more in line with previous academic work consisting of 7.27M PE files from the same sources as the baseline dataset, but collected over a much shorter time frame of July 2016 to November 2016.  No sampling was applied to the small dataset and its proportion of malware was 50\%.  For testing classification performance, we use the complete feed of PE files provided by VirusTotal and ReversingLabs from June 1, 2018 to August 31, 2018, which contains a total of 16.55M unique PEs with a 50/50 split of malware and goodware.

We use two small-scale datasets to perform our feature analysis.  The first is a random sample of 4,000 PE files taken from our baseline dataset with an equal split between malware and goodware, which we use to understand broad activation trends.  The second is a set of six artifacts from the NotPetya, WannaCry, and BadRabbit ransomware families, including loaders, payloads, and encryptors. We use this ransomware dataset to extract specific, concrete features for examination.

Our datasets are designed to follow best practices for malware classification research presented by Pendlebury et al. \cite{tesseract}, including time-based train/test splits, realistic class imbalances, and stratified sampling of malware family to reduce biases.  Our validation and testing sets contain samples that are strictly newer than the training set used to train our models.  For our baseline dataset, we implement an 80/20 goodware to malware ratio to capture the realistic assumption that the majority of files encountered by malware classifiers on user machines are, in fact, benign \cite{yen}.  Our baseline dataset also used stratified sampling to ensure that no single malware family is over-represented in our data.  As a final note, no special effort was made to filter our data sources, and as a result the dataset contains packed and otherwise obfuscated PE files, in addition to standard binaries.

%\begin{table}[h]
%\scriptsize
%\begin{center}
%\caption{Training details and results for three byte-based malware classification models on our test dataset.}
%\begin{tabular}{c|cc|cc|cl}
%\multicolumn{1}{l|}{} & \multicolumn{2}{c|}{\textbf{Dataset}}                  &	\multicolumn{2}{c|}{\textbf{Loss}} & \multicolumn{1}{l}{} &              \\
%\textbf{Model}        & \textbf{Size} & \multicolumn{1}{l|}{\textbf{Mal:Good}} & \textbf{Train} & \textbf{Valid} & \textbf{F1}          & \textbf{AUC} \\ \hline
%Small                 & 7.27M         & 50:50      &  0.023 & 0.069                            & 0.943                & 0.98         \\
%Baseline                 & 15.62M        & 20:80    &  0.003  & 0.273                           & 0.919                & 0.96         \\
%Baseline+Dropout         & 15.62M        & 20:80  & 0.022 & 0.745                         & 0.869                & 0.87        
%\end{tabular}
%\label{tbl:model_results}
%\end{center}
%\end{table}

\begin{figure*}[t!]
\centering
%\begin{center}
\begin{subfigure}[t]{0.33\textwidth}
\centering
\includegraphics[width=\textwidth, valign=t]{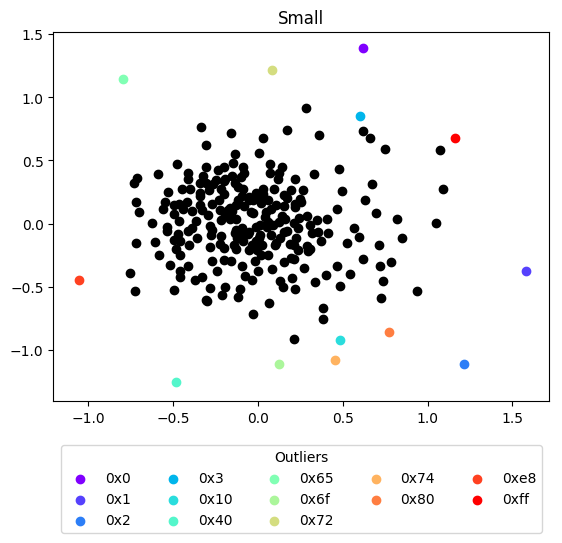}
\label{fig:embed:small}
\end{subfigure}\hfill
\begin{subfigure}[t]{0.33\textwidth}
\centering
\includegraphics[width=\textwidth, valign=t]{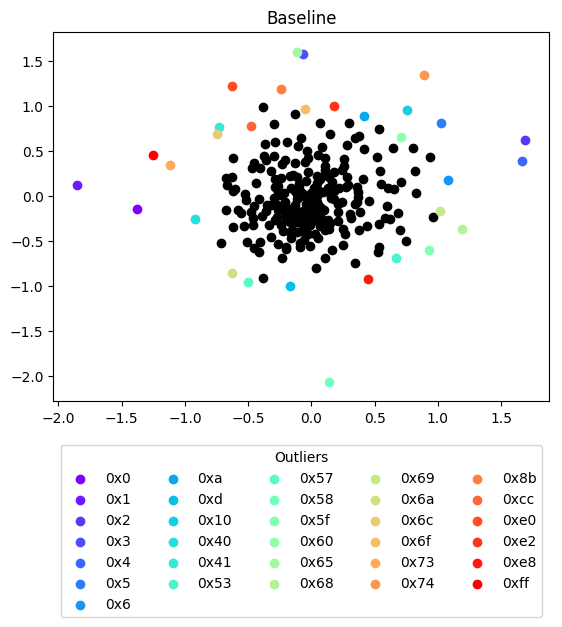}
\label{fig:embed:baseline}
\end{subfigure}\hfill
\begin{subfigure}[t]{0.33\textwidth}
\centering
\includegraphics[width=\textwidth, valign=t]{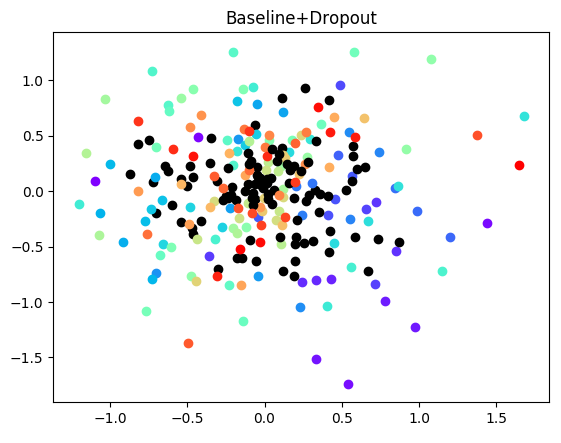}
\label{fig:embed:dropout}
\end{subfigure}
\caption{Visualization of embedding layers for byte-based malware classifiers using MDS.  Colored dots represent outliers discovered using HDBSCAN. Outliers not shown for Baseline+Dropout due to space restrictions.}
\label{fig:embedding}
\end{figure*}

\begin{table}[h]
\begin{center}
\caption{Training details and results for three byte-based malware classification models on our test dataset.}
\begin{tabular}{c|cc|cl}
\multicolumn{1}{l|}{} & \multicolumn{2}{c|}{\textbf{Train Data}}                 & \multicolumn{2}{c}{\textbf{Test Results}}               \\
\textbf{Model}        & \textbf{Size} & \multicolumn{1}{l|}{\textbf{Mal:Good}}  & \textbf{F1}          & \textbf{AUC} \\ \hline
Small                 & 7.27M         & 50:50      & 0.943                & 0.98         \\
Baseline                 & 15.62M        & 20:80     & 0.919                & 0.96         \\
Baseline+Dropout         & 15.62M        & 20:80  & 0.869                & 0.87        
\end{tabular}
\label{tbl:model_results}
\end{center}
\end{table}

\subsection{Methodology}
\label{sec:background:method}
We train three CNN models using the above architecture and data:  (1) a baseline model using the baseline training dataset, (2) a small model using the small dataset, and (3) a baseline+dropout model using the baseline data with the addition of dropout layers added according to recommendations by Srivastava et al. \cite{srivastava2014dropout} (with dropout rates of 0.1, 0.25, 0.25, 0.25, 0.5, and 0.5).  All models were initialized with Xavier initialization, and trained for ten epochs using the Momentum optimizer with decay.  Training and validation losses for all models converged before the end of the tenth epoch.

The results of our initial evaluation of each model on our test dataset are shown in Table \ref{tbl:model_results}.  In general, the performance of the small and baseline models falls in line with the results reported in previous work, while the model using dropout performs significantly worse.  Notice that, paradoxically, the small model performs the best despite having significantly older and fewer data points to train on.  This result is at odds with common wisdom that more data and improved regularization will result in better generalization and, consequently, better classification performance.  We will explore the reasons for this in the coming sections by comparing the learned features from each model.

\begin{figure*}[t!]
\centering
%\begin{center}
\begin{subfigure}[t]{0.33\textwidth}
\centering
\includegraphics[width=\textwidth, valign=t]{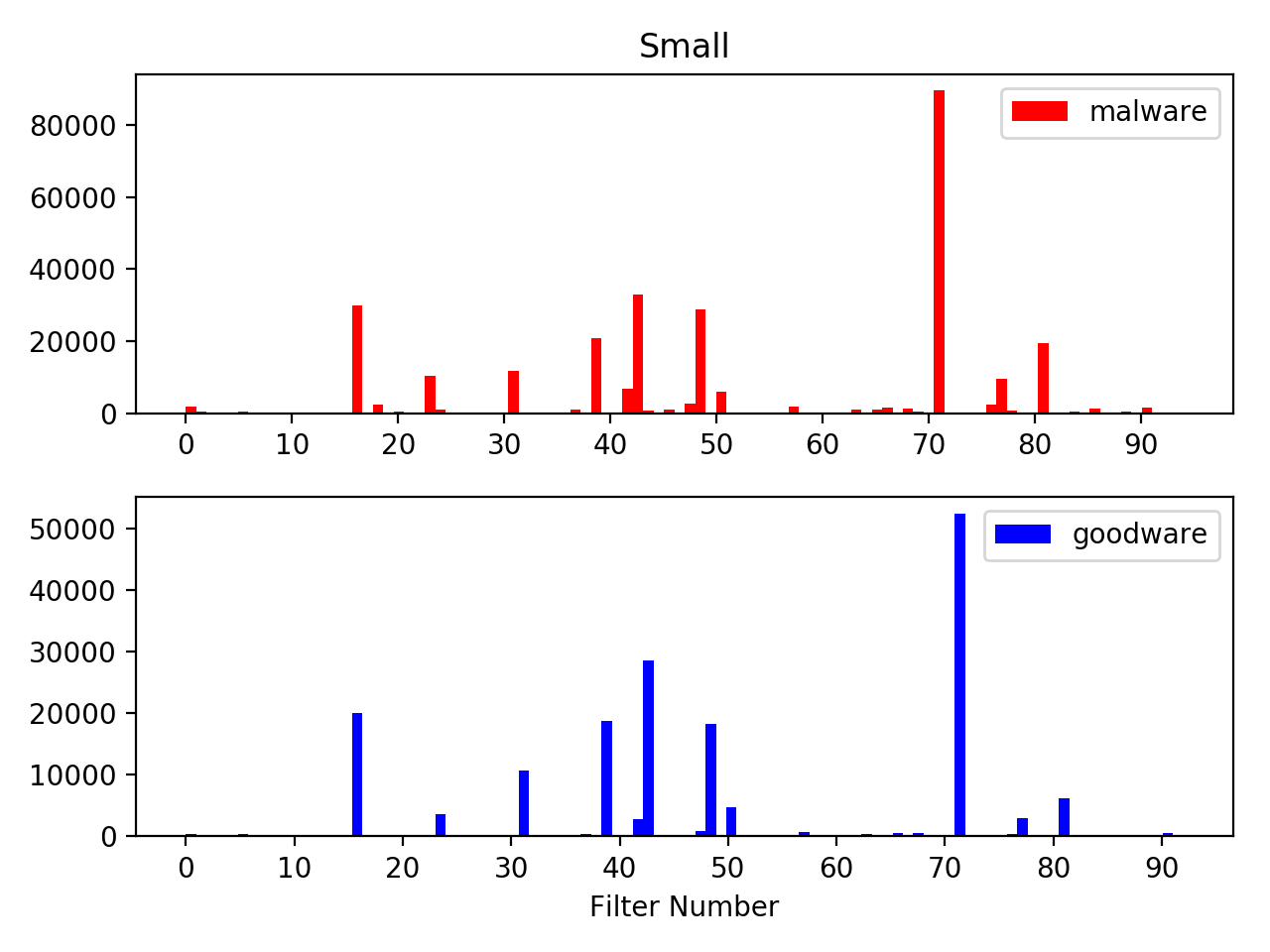}
%\caption{Small}
\label{fig:filter:small}
\end{subfigure}\hfill
\begin{subfigure}[t]{0.33\textwidth}
\centering
\includegraphics[width=\textwidth, valign=t]{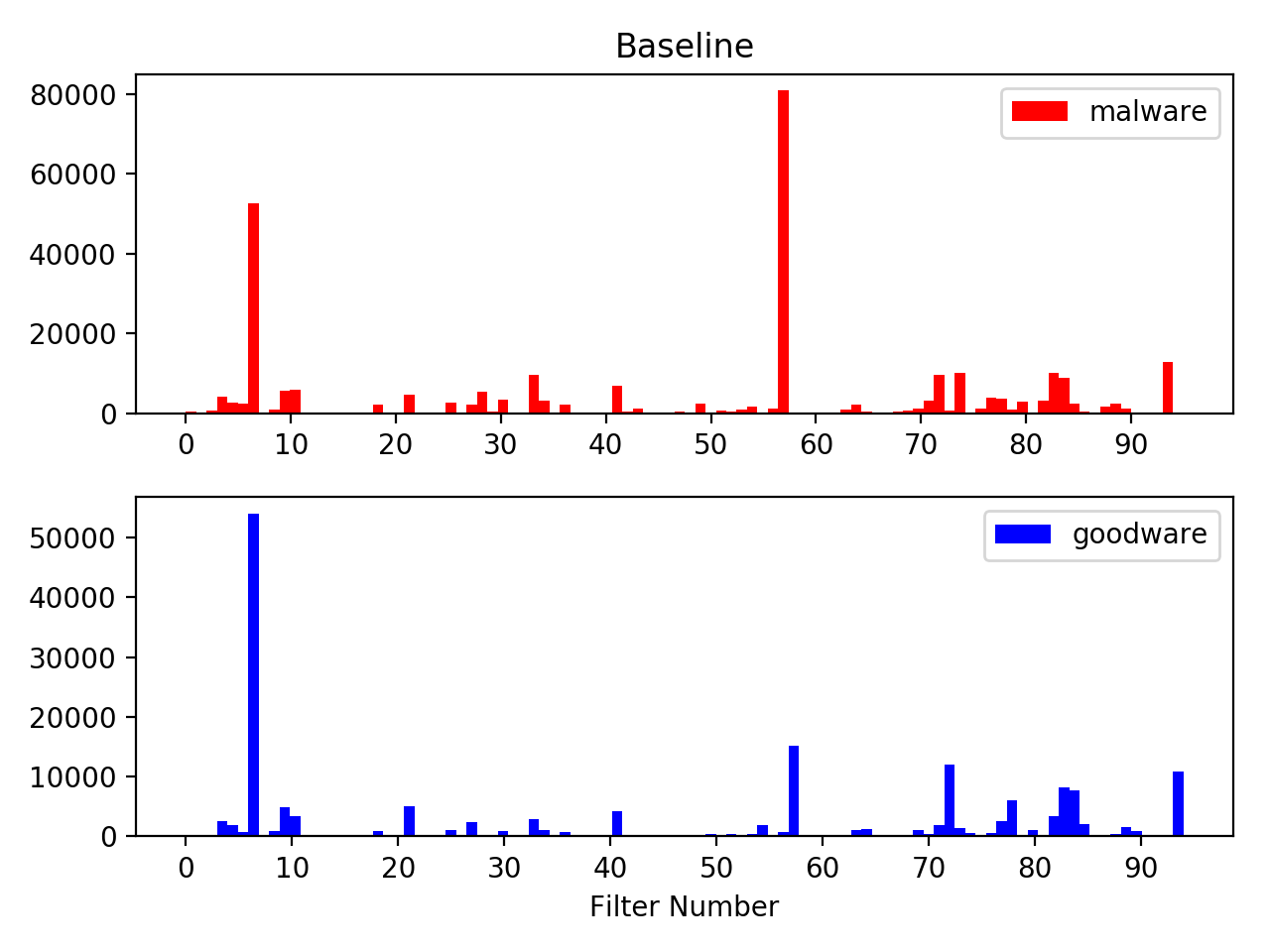}
%\caption{Baseline}
\label{fig:filter:baseline}
\end{subfigure}\hfill
\begin{subfigure}[t]{0.33\textwidth}
\centering
\includegraphics[width=\textwidth, valign=t]{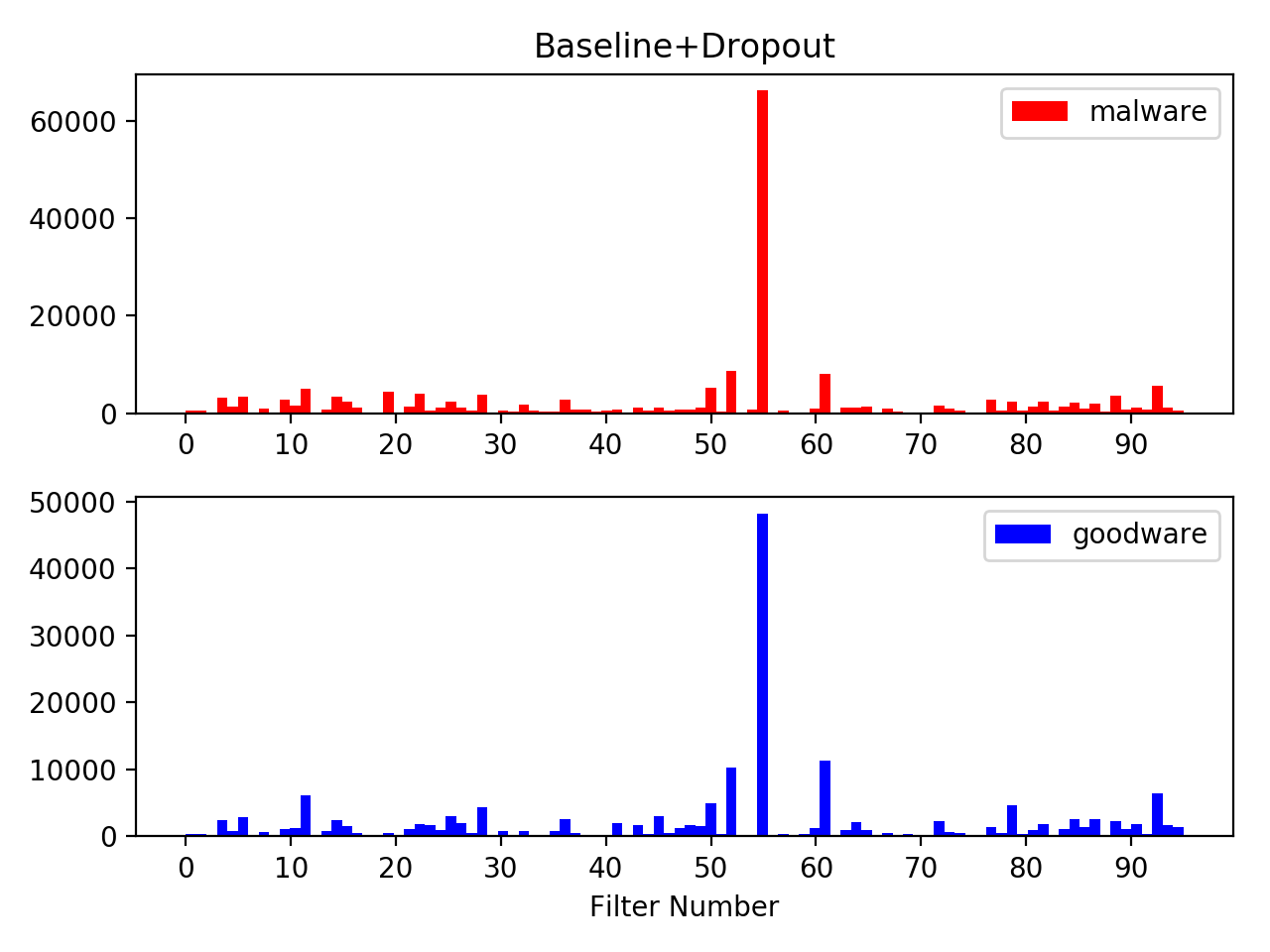}
%\caption{Baseline+Dropout}
\label{fig:filter:dropout}
\end{subfigure}
\caption{Distribution of top-100 activations across 96 first-level filters.}
\label{fig:filters}
\end{figure*}

\begin{figure*}[t!]
\centering
%\begin{center}
\begin{subfigure}[t]{0.33\textwidth}
\centering
\includegraphics[width=\textwidth, valign=t]{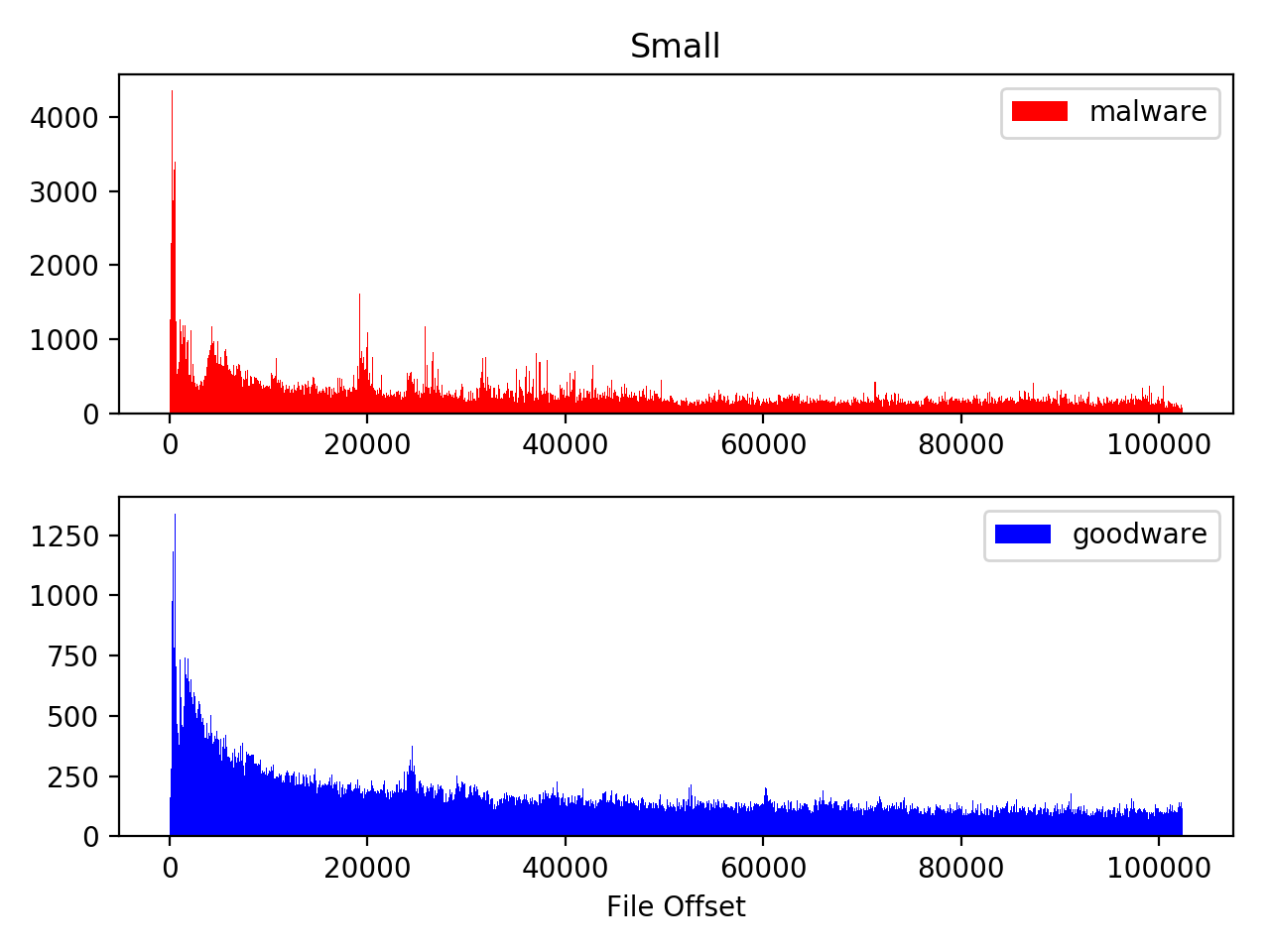}
%\caption{Small}
\label{fig:filter:small}
\end{subfigure}\hfill
\begin{subfigure}[t]{0.33\textwidth}
\centering
\includegraphics[width=\textwidth, valign=t]{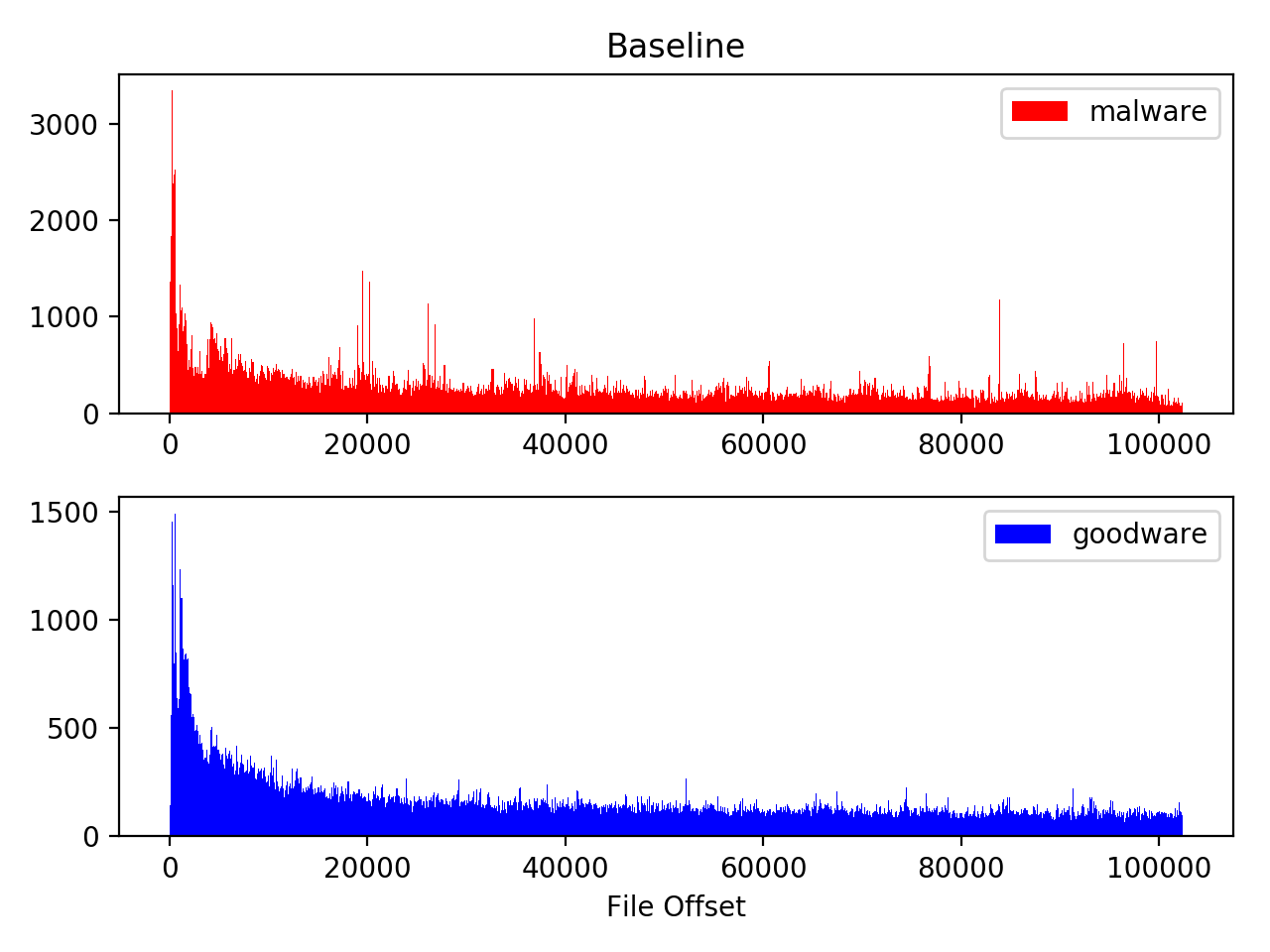}
%\caption{Baseline}
\label{fig:filter:baseline}
\end{subfigure}\hfill
\begin{subfigure}[t]{0.33\textwidth}
\centering
\includegraphics[width=\textwidth, valign=t]{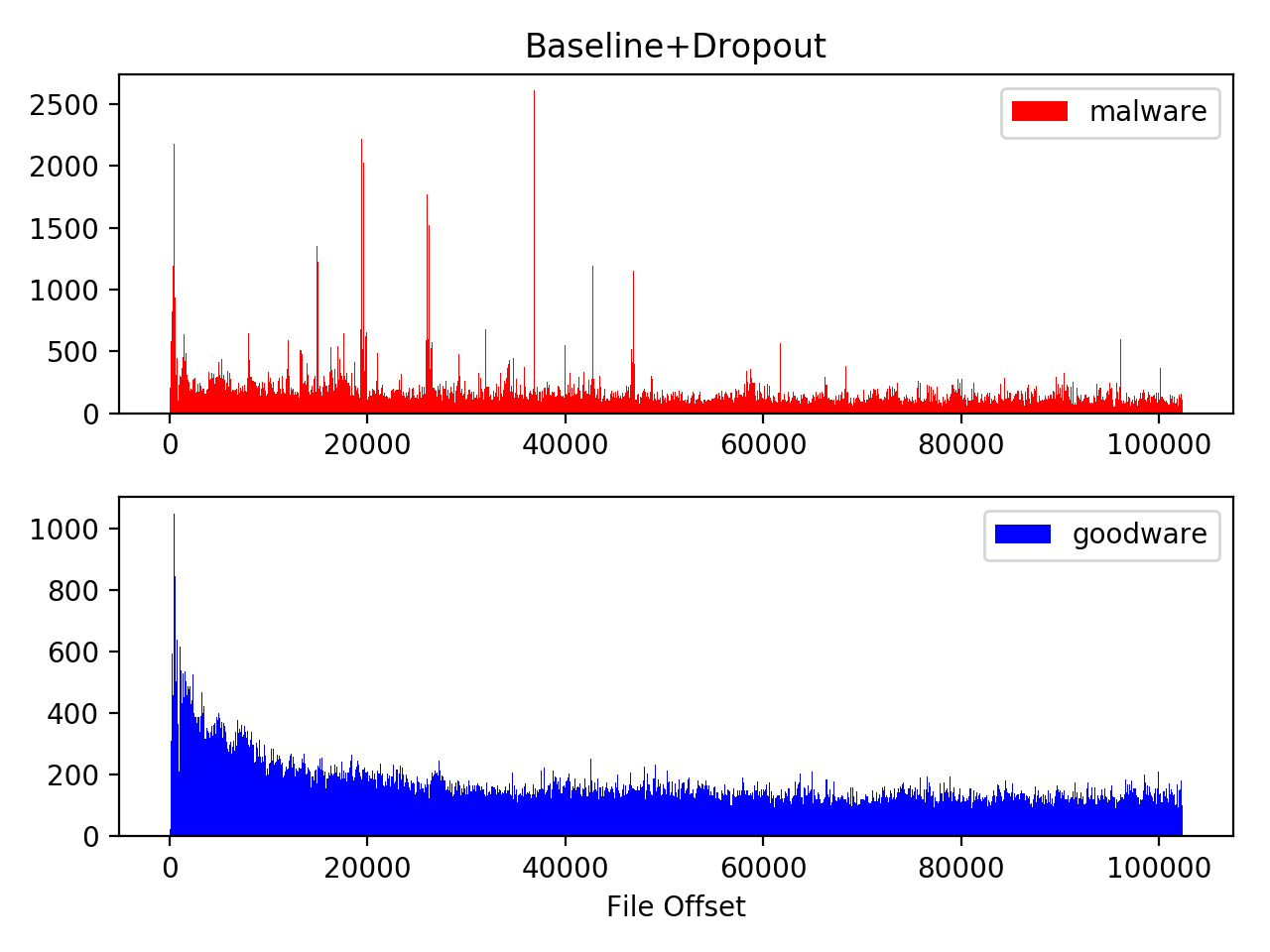}
%\caption{Baseline+Dropout}
\label{fig:filter:dropout}
\end{subfigure}
\caption{Distribution of top-100 activation locations across file offsets.}
\label{fig:filter}
\end{figure*}

% 16,553,539

%\vfill\null

\section{Embedding Layer}
\label{sec:embedding}

Here, we start our exploration by comparing the embedding layers of our three models.  Since the embedding matrix is treated as a set of learnable parameters during training, the locations of each of the bytes (plus padding symbol) can be manipulated within the 10-dimensional embedding space. Intuitively, if one or more bytes can be used interchangeably within the learned features (particularly those of low-level convolutional filters), then they will be densely clustered within the embedding space.  Conversely, outliers within this space indicate byte values that are distinguished in some way and incur significant cost to swap.

To identify these clusters and their outliers, we apply hierarchical density-based clustering (HDBSCAN) \cite{campello2013hdbscan} on the points in the embedding space.  As the name suggests, this algorithm extracts clusters based on their density, but unlike traditional DBSCAN does not require near-uniform density among clusters since the appropriate distance (i.e., $\epsilon$) is inferred for each neighborhood from the data, in a hierarchical manner.  The two parameters that we control are (1) the minimum number of points necessary to define a cluster, which we set to two, and (2) the minimum number of samples to define a neighborhood used to calculate cluster density, which we set to one.  Both parameters settings allow for a fairly liberal and relaxed density calculation and should maximize the number of clusters (resp., minimize outliers).  We visualize the embedding space, as shown in Figure \ref{fig:embedding}, using metric multi-dimensional scaling (MDS) \cite{borg1997mds}, which we found provides a good approximation for the underlying clustering behavior.

A few broad trends are immediately evident when comparing the embedding space visualizations and discovered outliers -- the number of outliers and overall sparsity of the space increases as we move from our small model to the regularized baseline+dropout model.  The small model has 13 outliers, which increases to 31 for baseline, and up to 158 for baseline+dropout. Both the small and baseline model had a small number of clusters, with one of those clusters containing the vast majority of bytes, indicating that many of the bytes are interchangeable within higher-level features.  At the same time, the increase in sparsity among the models and large numbers of outliers for the baseline+dropout model could be indicative of the models moving toward more restrictive byte sequence features in the low-level convolutional filters.

Additionally, the outliers shared by the small and baseline models are quite informative of what features might be learned at the upper layers of the model.  These outliers include typical x86 register values (eax/0x0, ecx/0x1, edx/0x2, ebx/0x3), several ASCII characters (`e'/0x65, `o'/0x6f, `t'/0x74), and the 'call' instruction (0xe8).  Notably, the baseline model also includes several more ASCII characters (`W'/0x57, `s'/0x73, `\textbackslash n'/0xa) and instructions (loop/0xe0, loopne/0xe2), which again points to increasing specificity in the learned features.  More generally, however, we see a clear predilection across all models toward features associated with ASCII strings and control flow instructions.  As we will see in the next section, these outlier bytes will play a key role in defining the low-level features, such as import names and instruction sequences.

\section{Low-Level Features}
\label{sec:low-level}
With the results of our embedding analysis in mind, we now move up one level to the first convolutional layer.  In our architecture, this layer has 96 filters with a kernel width of eleven bytes and a stride of one, resulting in 102,390 convolutions over our embedded 100KB input sequence.  We examine these convolutions in two ways.  First, we look for broad trends in the locations and filter usage for the three models, along with how those trends differ between malware and goodware classes.  Second, we look at the specific features identified by the models in our ransomware dataset by parsing the binaries using PEFile \cite{pefile} and disassembling them with BinaryNinja \cite{binaryninja}.

To begin, let us draw some insight into the differences in filter usage and activation locations across our models by extracting the top-100 activations from each of our 4,000 PE file samples.  Figure \ref{fig:filters} shows the distribution of these activations across our 96 first-level filters.  One interesting trend that arises is that utilization of filters increases dramatically as we go from our small model to the baseline+dropout model.  At the same time, the vast majority of activations remain on a single filter while the remainder become more diffusely spread among the other filters.  While the filters used in the malware and goodware classes remains mostly the same, the number of malware activations is significantly higher than goodware for the small and baseline models, but equalizes for the baseline+dropout model.  More concretely, the filters in the small and baseline models have five times more malware activations than the baseline+dropout model, which itself shows only a nominal (55\% vs. 45\%) bias toward malware activations.  This is an interesting observation because it hints at the idea that these low-level features are, in some way, more biased toward malware even though the underlying dataset may have an imbalance toward goodware -- the small model has 50\% goodware, while baseline has 80\% goodware.  By adding regularization in the form of dropout, the model appears to be learning more generically-applicable features across the two classes, but at the cost of worse overall classification performance.

We also examine the location of the top activations by file offset, as illustrated in Figure \ref{fig:filter}.  Again, we observe some distinctive trends as we increase dataset size and regularization, but only in our malware class.  In particular, the location of activations moves from the start of the files (i.e., PE headers) in the small model and gets distributed throughout the file as we move toward the baseline+dropout model.  Most notably, the baseline+dropout model no longer focuses the majority of activations on the PE headers, but instead spreads those activations around the same high-activation areas found in other models.  That is, the areas of interest identified by the models remains relatively stable, but their overall impact on the low-level activations is markedly different.  For goodware, on the other hand, the overall distribution looks quite similar across all models, albeit with less concentration toward the start of the files in the baseline+dropout model.  Taken together with the filter analysis above, we see the impact of regularization as a trend of moving away from heavily-weighted, malware-oriented features and toward more generic features with more uniform activation values.

\begin{table}[h]
\scriptsize
\caption{Example features from disassembled ransomware.}
\begin{tabular}{c|cc}
\multicolumn{1}{l|}{}                      & \multicolumn{2}{c}{\textbf{Features}}                                                                                                                                                                                                                                                                                                                                                                   \\
\multicolumn{1}{c|}{\textbf{Model}}                      & \textbf{Strings}                                                                                                                                                                   & \textbf{Instructions}                                                                                                                                                                                              \\ \cline{1-3} 
\multirow{2}{*}{\textbf{Small}}            & \textbf{Filter 71: `C', `r', `@'}                                                                                                                                                  & \textbf{Filter 16: Push sequences}                                                                                                                                                                                 \\
                                           & \multicolumn{1}{l}{\begin{tabular}[c]{@{}l@{}}(0x40f0c8L): tGenKey.\\ (0x40f0d0L): CryptDec\\ (0x40f0d8L): rypt....\\ (0x40f0e0L): CryptEnc\\ (0x40f0e8L): rypt....\end{tabular}}  & \multicolumn{1}{l}{\begin{tabular}[c]{@{}l@{}}(0x10007edbL): je,0x10007ff1\\ (0x10007ee1L): push,0xff\\ (0x10007ee6L): push,edi\\ (0x10007ee7L): push,0x10007ca5\\ (0x10007eecL): push,0x4\end{tabular}}           \\ \cline{2-3} 
\multirow{2}{*}{\textbf{Baseline}}         & \textbf{Filter 83: `r', `s'}                                                                                                                                                       & \textbf{Filter 57: Function calls}                                                                                                                                                                                 \\
                                           & \multicolumn{1}{l}{\begin{tabular}[c]{@{}l@{}}(0x40d850L): ....GetP\\ (0x40d858L): rocAddre\\ (0x40d860L): ss..R.Lo\\ (0x40d868L): adLibrar\\ (0x40d870L): yA....Gl\end{tabular}}  & \multicolumn{1}{l}{\begin{tabular}[c]{@{}l@{}}(0x4046b4L): push,0x0\\ (0x4046b6L): push,0x0\\ (0x4046b8L): push,0x1\\ (0x4046baL): push,0x0\\ (0x4046bcL): call,dword,15042\end{tabular}}                          \\ \cline{2-3} 
\multirow{2}{*}{\textbf{Baseline+Dropout}} & \textbf{Filter 11: `Directory'}                                                                                                                                                    & \textbf{Filter 61: mov sequences}                                                                                                                                                                                  \\
                                           & \multicolumn{1}{l}{\begin{tabular}[c]{@{}l@{}}(0x40d9e0L): ctoryW..\\ (0x40d9e8L): N.Create\\ (0x40d9f0L): ,Director\\ (0x40d9f8L): yW....Ge\\ (0x40da00L): tTempPat\end{tabular}} & \multicolumn{1}{l}{\begin{tabular}[c]{@{}l@{}}(0x408d65L): je, 0x408d6a\\ (0x408d67L): mov, dword , edx\\ (0x408d6aL): mov, esi, dword\\ (0x408d6dL): mov, dword, esi\\ (0x408d70L): mov, ecx, dword\end{tabular}}
\end{tabular}
\label{tbl:features}
\vspace{-1em}
\end{table}

To further substantiate our intuitions about low-level features, we now compare the disassembly of the activation locations for several of the most prevalent filters applied to our ransomware dataset.  Doing so provides both a notion of the types of features identified, as well as how those features vary across our models.  Generally, the filters appear to have two primary types of features: common instruction sequences and ASCII strings (e.g., import names).  Some example features for each type are shown in Table \ref{tbl:features}.  While all filter activations have a certain level of variability due to incidental detection of the relatively short eleven-byte sequences, there is a clear trend toward more precise, limited features in our baseline and baseline+dropout models versus the small model.  That is, filters for the baseline+dropout model are precisely constrained to very specific strings and instruction sequences, whereas the small model has relatively loose filters resulting in more spurious activations.  For example, Table \ref{tbl:features} shows import activations for the small and baseline models where the commonality is based on the number of specific ASCII characters in the convolution, thereby allowing activations for a large class of similar imports, while the baseline+dropout model has entire filters looking for only a single string, such as ``Directory."  Similarly, the most prevalent filter in the baseline+dropout model, filter 55, activates solely on areas of padding between sections containing multiple padding bytes (0xff).  We observe similar behaviors among the instruction sequences, where increasingly specific context restrictions are imposed, from an activation containing any push instruction in the small model to a specific sequence of mov instructions in the case of the baseline+dropout model.

Overall, these results highlight some very important properties of our models.  For one, we see the relationship between the outlier bytes identified in our embedding analysis and their role as keys or anchors used to define many of the most important filter features in the first convolutional layer.  There is also certainly a heavy reliance on import name features, with that reliance increasing in the baseline+dropout model.  Additionally, the instruction sequences coincidentally capture some useful behavioral features, such as the use of push-call sequences to fingerprint certain types of function calls (e.g., Windows APIs with multiple parameters).  Together, the import name and instruction sequence features represent two sides of the same underlying import usage behavior that is often included in manually-derived features, though the connection between the import and where (or if) it is actually used is clearly lost due to the inability of the CNN architecture to understand the offsets and indirection used.  Finally, we begin to see clues that the poor performance of the baseline+dropout model may be due to the reduction in importance for malware-oriented features, which again challenges our natural intuition that increased generalization should result in improved classification performance.

\begin{figure*}[t!]
\centering
\footnotesize
%\begin{center}
\begin{subfigure}[t]{0.33\textwidth}
\centering
\includegraphics[width=\textwidth, valign=t]{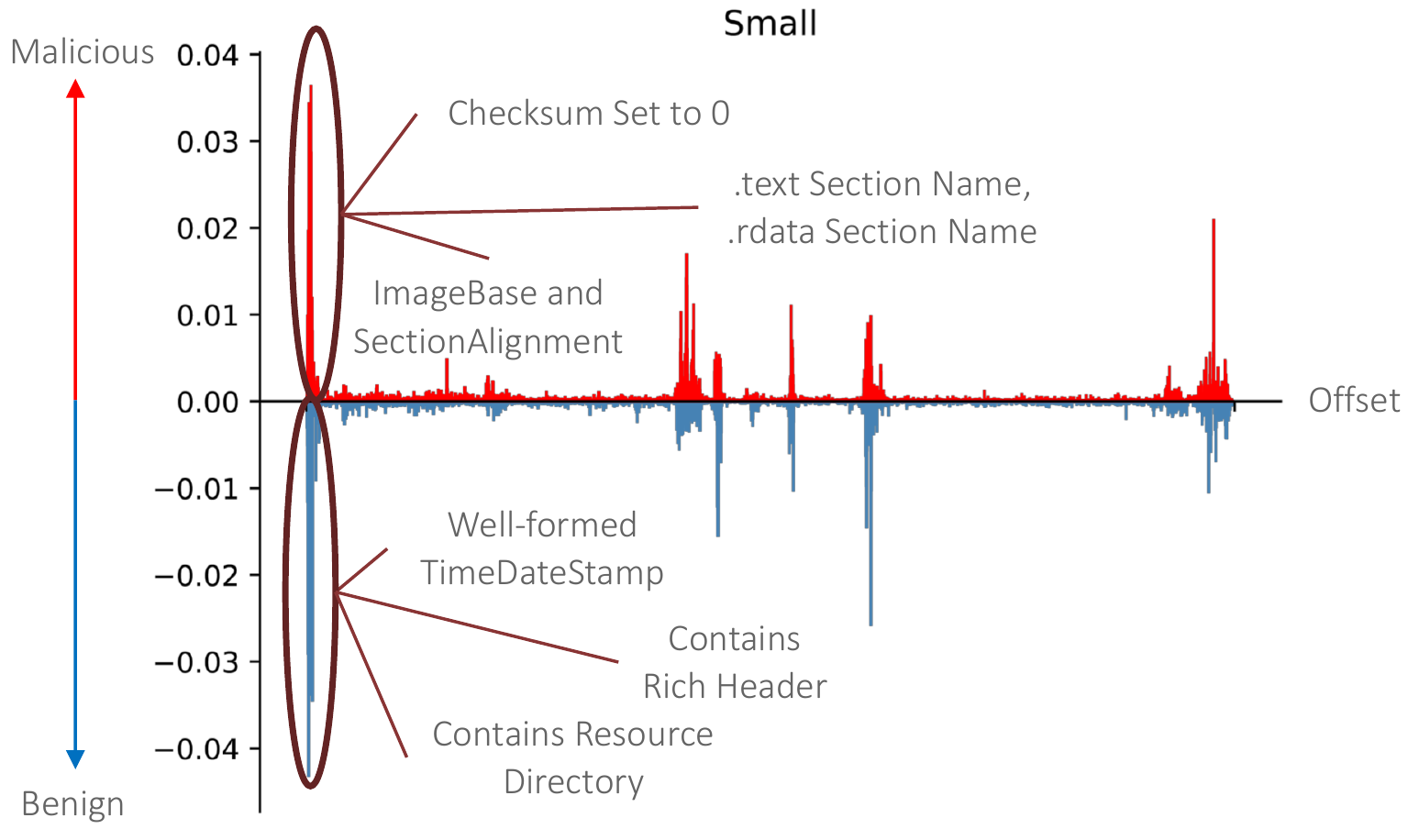}
%\caption{Small}
\label{fig:shap:small}
\end{subfigure}\hfill
\begin{subfigure}[t]{0.33\textwidth}
\centering
\includegraphics[width=\textwidth, valign=t]{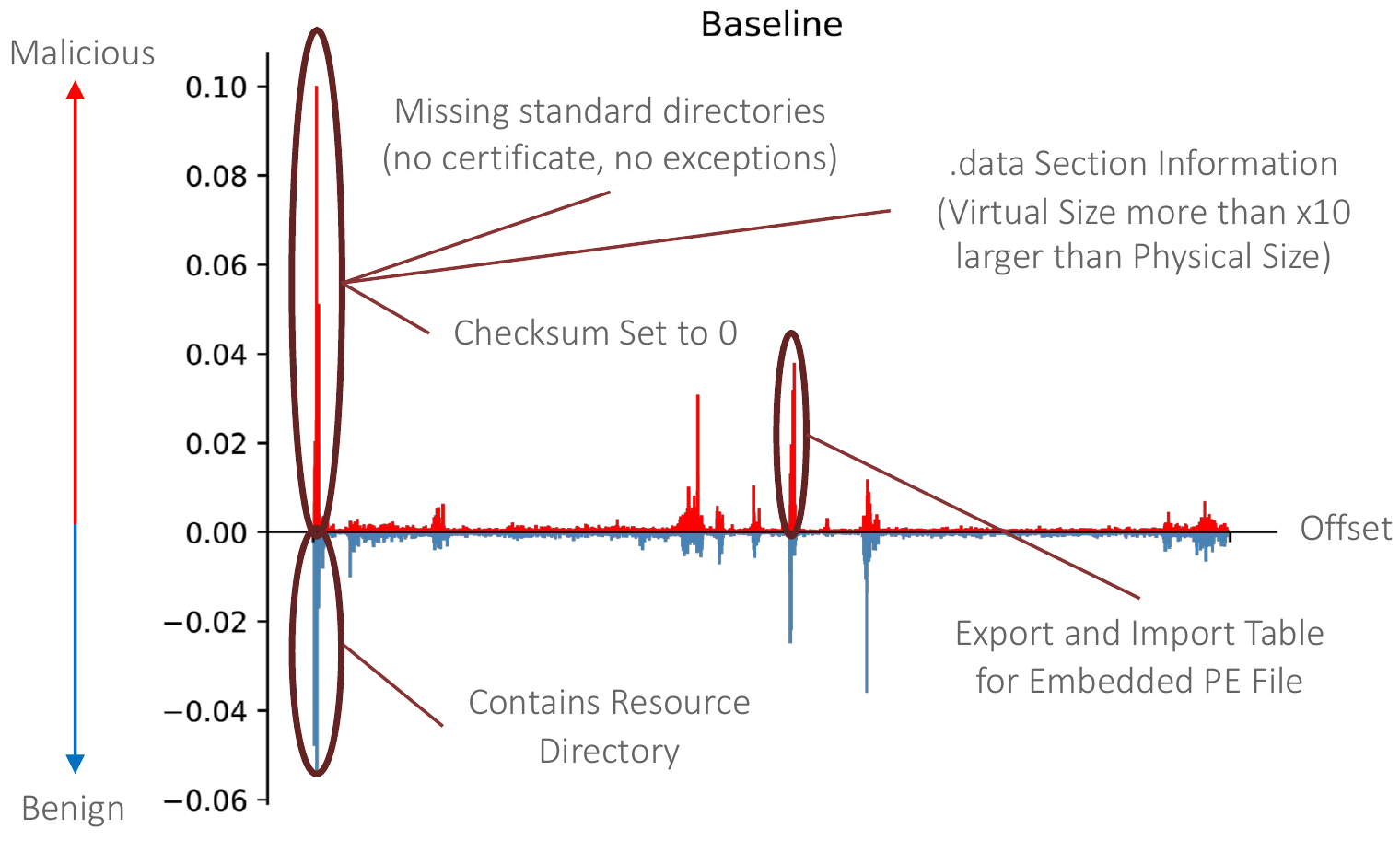}
%\caption{Baseline}
\label{fig:shap:baseline}
\end{subfigure}\hfill
\begin{subfigure}[t]{0.33\textwidth}
\centering
\includegraphics[width=\textwidth, valign=t]{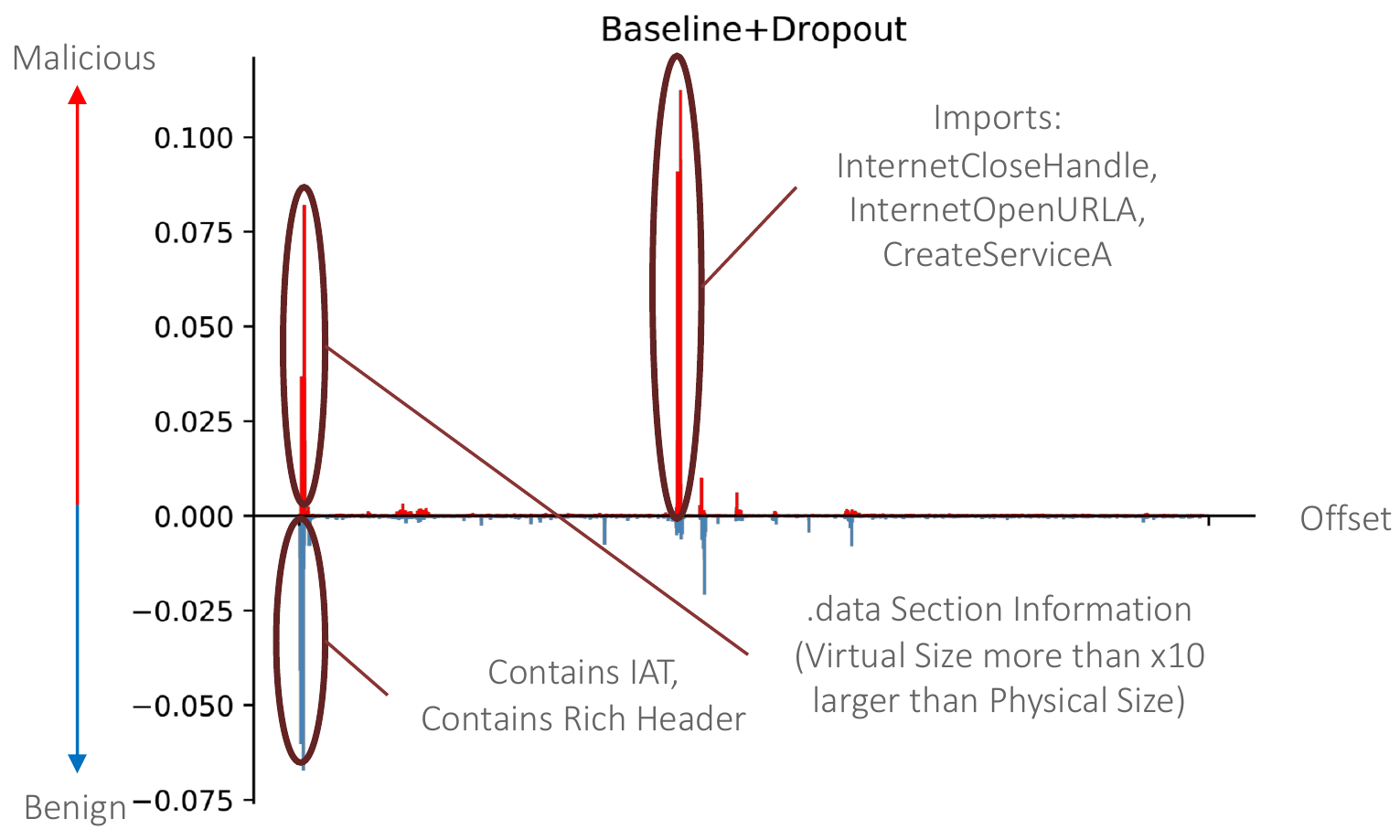}
%\caption{Baseline+Dropout}
\label{fig:shap:dropout}
\end{subfigure}
\caption{Distribution of SHAP values for the WannaCry worm. SHAP values greater than zero indicate maliciously-oriented features. The largest malicious and benign SHAP values are annotated with their underlying semantics in the disassembly.}
\label{fig:shap}
\end{figure*}

\section{End-to-End Features}
\label{sec:end-to-end}
For our final set of experiments, we take a holistic view on the learned features by connecting segments of the input sequence with their contribution to the final score produced by the model.  We use the SHapley Additive exPlanations (SHAP) framework proposed by Lundberg and Lee \cite{lundberg2017shap} to assign contributions to each input feature -- in our case individual bytes -- using concepts derived from cooperative game theory (i.e., Shapley value).  Specifically, we use the GradientSHAP \cite{gradientshap} method, which combines the Integrated Gradients \cite{sundararajan2017integrated} and SmoothGrad \cite{smilkov2017smoothgrad} techniques with the SHAP framework.  Intuitively, the method calculates the expectation of the gradients between the given input and a large number of randomly sampled feature vectors generated from a background dataset.  The end result is a precise attribution of the input features that contributed to a given classification.  Our experiments used the 4,000 sample analysis dataset as the background set, 1,000 random samples, and no local smoothing when computing SHAP values.  We also note that GradientSHAP requires a fully differentiable model, which is incompatible with our embedding layer.  Instead we perform the explanation after embedding and recover each byte's SHAP value by taking their sum over the embedding dimensions.

We apply GradientSHAP to each ransomware artifact in our analysis dataset and, like our low-level filter analysis, reference the disassembly of the files to connect the byte-oriented contributions to human-understandable features.  To make the analysis of features easier, we combined contiguous bytes with the same contribution directionality (i.e., toward malware or goodware) into segments and define the segment SHAP value as the sum of the constituent byte SHAP values.  Figure \ref{fig:shap} shows these segment SHAP values for the WannaCry worm (SHA1: db349b97c37d22f5ea1d1841e3c89eb4).  Positive SHAP values indicate that the bytes in the segment are pushing the model toward a classification of malware, while negative values indicate a contribution toward a goodware verdict.

Starting again with broad observations across the three models, we see that the increasing specificity of features in the baseline and baseline+dropout models is evident even in our end-to-end analysis.  With the baseline+dropout model in particular, the evidence for malware classification is concentrated on a small number of segments related to only a few underlying structures in the PE file.  By contrast, the small model has evidence spread throughout the byte sequence, including segments that overlap with the other two models.  At the same time, the small model also includes a large number of spurious benign features that significantly reduce the malware probability for the file.

Diving deeper, we see a number of important (i.e., high SHAP value) features annotated in the Figure \ref{fig:shap} that clearly align with features typically found via manual feature engineering.  One such feature is the detection of a checksum set to zero in the small and baseline models.  While not an inherently malicious property of the file, a large fraction of malware contain incorrect checksums, which makes it a useful distinguishing feature.  Interestingly, in the baseline+dropout model, the checksum-related feature actually flips and becomes a goodware-oriented feature that checks for a non-zero value -- reinforcing our findings that regularization forces the model away from malware-centric features.  Other features of note include detecting the absence of the standard security directory that exists with signed code, identifying import and export tables within an embedded PE file, and several import names.  We hypothesize that these complex, high-level features are learned based on the fact that structures found later in the PE file are shifted, which creates unusual locality among lower-level features that the model is able to take advantage of due to its  hierarchical pooling architecture.

The results of our analysis on the ransomware artifacts also contained one high-level feature whose role in pushing the model towards benign classifications was not obvious to us at first glance.  Specifically, we identified several instances where the most important benign end-to-end features referenced the so-called Rich header \cite{richheader}, as annotated in Figure \ref{fig:shap}.  The Rich header is added only by Microsoft's linker and contains an XOR encrypted set of linker metadata about the binary, such as the number of objects and linker version used.  However, since this information is effectively randomized we should not expect the model to learn useful information from this area of the binary.  What we believe is happening is that the model is, like other high-level structures, learning whether the Rich header is present based on the location of a string that directly follows the Rich header.  If the Rich header is detected, then the model knows that a Microsoft linker is used and, as it turns out, a non-trivial fraction of malware is built using non-Microsoft toolchains (e.g., Delphi) \cite{ciscodelphi, fireeyedelphi}.  The linker and its version are actually common manually-derived features, and the model appears to have effectively inferred that same feature via the complex interplay of several low-level features, demonstrating the potential of byte-based models to extract unusual but highly-effective features without human intervention.

\section{Conclusion}
\label{sec:conclusion}
Byte-based deep learning classifiers have proven to be a viable alternative to traditional machine learning for malware classification, without the cost of manual feature engineering.  Through our experiments, we have explored what the deep neural networks are learning about Windows PE files in order to separate malware from goodware.

Our findings show that import-related features play a very important role in the operation of the models -- appearing at all levels of our analysis, from embedding to end-to-end features.  We also discovered that the embedding layer and low-level convolutional filters learned features related to ASCII strings and instruction sequences, with some filters capturing useful behavioral indicators like the use of system calls through push-call sequences.  However, these instruction-oriented features do not appear in our end-to-end analysis.  Instead, the depth and complexity of the model's architecture allowed it to learn intuitive and meaningful features, including incorrect checksums and the presence of certificates.  The models were also able to learn some novel features, such as the presence of the Rich header as a proxy for linker type information.  Many of these features are used in machine learning classifiers with manually-derived features (e.g., Saxe and Berlin \cite{saxe2015deep}).

Perhaps the most interesting result was the paradoxical impact that increased data volume and regularization had on classification performance.  Although the highly-regularized dropout model learned more general features across both malware and goodware classes, its performance was substantially worse than a model with an older and smaller training set.  One explanation for this behavior is that, malware-oriented features (or the lack thereof) are strong indicators of class membership for both malware and goodware -- similar to the way traditional antivirus signatures operate.  Another potential explanation is that the baseline+dropout model was not trained for long enough to fully converge, which Srivastava et al. warn could take 2-3 times longer than the original architecture \cite{srivastava2014dropout}.  However, we note that validation loss had plateaued before the final training epoch, which makes that explanation somewhat less likely.

To underscore the importance of these findings in identifying promising avenues of future work, we can compare our results to those of Demetrio et al., who performed a similar analysis on the MalConv byte-based malware classifier \cite{luca2019explaining}.  In their analysis, Demetrio et al. found that nearly all of the most influential activations for the MalConv model occurred in the headers of the PE file due to its use of a single, gated convolutional layer and global pooling.  Overall, the MalConv features were found to be mostly devoid of semantic meaning.  These results stand in stark contrast to our own, which show that with sufficient model depth and training data volume, deep neural networks can learn feature representations that closely mimic those created by subject-matter experts.  Moreover, our results indicate that influential features are learned throughout the entire length of the input byte sequence, thereby providing evidence that the type of hierarchical architecture examined here may learn more diverse and robust features.  Taken as a whole, our results provide interesting insight into the role that architecture, regularization, and data characteristics play in developing byte-based malware classifiers, and hopefully improve our overall understanding of the malware classification problem.

\vfill

\bibliographystyle{ieeetran}
\bibliography{activation}

\end{document}